\documentclass[11pt]{article}

\usepackage{acl}

\usepackage{times}
\usepackage{latexsym}
\usepackage[T1]{fontenc}
\usepackage[utf8]{inputenc}
\usepackage{microtype}
\usepackage{inconsolata}
\usepackage{graphicx}
\usepackage{booktabs}
\usepackage{multirow}
\usepackage{amsmath}
\usepackage{amssymb}
\usepackage{xcolor}
\usepackage{tikz}
\usepackage{placeins}
\usepackage{float}
\usetikzlibrary{arrows.meta,positioning,fit,calc}

\usepackage{fancyhdr}
\newcommand{\githublink}{https://research.kaysarulanas.me/} 
\title{SciRet: A Compute-Aware Empirical Study of Retrieval and Reranking for Scientific RAG}

\author{
  {\bf Kaysarul Anas Apurba}$^{1}$\thanks{~Corresponding author.} \and 
  {\bf Md. Hasibul Hasan}$^{1}$ \and 
  {\bf Rofiqul Alam Shehab}$^{2}$ \and
  {\bf Asab Azad}$^{1}$ \\
  $^{1}$Laurentian University \qquad $^{2}$North South University \\
  \texttt{kaysarulanas2@gmail.com} \\
  \url{https://github.com/anaskaysar/sciret}
}

\begin{document}
\maketitle
\thispagestyle{fancy}

\begin{abstract}
We introduce SciRet, a compute-aware empirical study of retrieval-augmented generation for scientific question answering over CORD-19. Rather than proposing a new model, we evaluate a fixed scientific RAG pipeline across three corpus scales: 1,034 chunks (1K papers), 5,160 chunks (5K papers), and 15,480 chunks (15K papers). The pipeline combines sentence-window chunking, BM25, BGE-M3 dense retrieval, reciprocal rank fusion, optional cross-encoder reranking, and grounded answer generation. Across these settings, hybrid retrieval is more robust than either sparse-only or dense-only retrieval in our setting, reaching Recall@10 of 1.000 at 1K and 15K. In contrast, an MS MARCO-trained cross-encoder reranker reduces precision on the scientific corpus, suggesting that domain mismatch can outweigh the benefits of stronger query-passage interaction. Generation faithfulness measured with RAGAS increases with corpus scale in our setup. Retrieval evaluation uses pseudo-relevance labels derived from the hybrid system, so we treat the results as controlled comparative evidence rather than a benchmark claim. We release code, indexes, and evaluation outputs to support replication and follow-up studies.
\end{abstract}

\section{Introduction}

Retrieval-augmented generation (RAG) is now a common architecture for knowledge-intensive question answering, but scientific literature remains a difficult setting. Scientific queries often depend on precise terminology, domain-specific evidence, and citation-grounded answers. A RAG pipeline that works well for web QA may not transfer cleanly to scientific text, especially when retrieval, reranking, and generation are trained or tuned on different domains.

This paper asks a focused empirical question: \emph{how do standard retrieval and reranking components behave when the same scientific RAG pipeline is evaluated across increasing corpus scale?} We study CORD-19, holding preprocessing, chunking, embedding model, retrieval settings, and generation settings fixed across 1K, 5K, and 15K paper samples. This controlled design isolates scale effects and makes failure modes easier to inspect.

\begin{figure*}[t]
\centering
\resizebox{\linewidth}{!}{%
\begin{tikzpicture}[
  font=\small,
  node distance=0.9cm and 1.05cm,
  box/.style={draw, rounded corners=2pt, align=center, minimum height=0.75cm, minimum width=2.3cm, fill=blue!5},
  proc/.style={draw, rounded corners=2pt, align=center, minimum height=0.75cm, minimum width=2.6cm, fill=teal!7},
  eval/.style={draw, rounded corners=2pt, align=center, minimum height=0.75cm, minimum width=2.6cm, fill=orange!10},
  arrow/.style={-{Latex[length=2mm]}, thick}
]
\node[box] (data) {CORD-19\\titles + abstracts};
\node[proc, right=of data] (chunk) {Sentence-window\\chunking};
\node[proc, above right=0.65cm and 1.1cm of chunk] (bm25) {BM25\\sparse index};
\node[proc, below right=0.65cm and 1.1cm of chunk] (dense) {BGE-M3\\dense index};
\node[proc, right=3.4cm of chunk] (rrf) {Reciprocal Rank\\Fusion};
\node[proc, right=of rrf] (rerank) {Optional MS MARCO\\cross-encoder};
\node[box, right=of rerank] (gen) {Grounded\\answer generation};
\node[eval, right=of gen] (metrics) {Retrieval + RAGAS\\evaluation};

\draw[arrow] (data) -- (chunk);
\draw[arrow] (chunk) |- (bm25);
\draw[arrow] (chunk) |- (dense);
\draw[arrow] (bm25) -| (rrf);
\draw[arrow] (dense) -| (rrf);
\draw[arrow] (rrf) -- (rerank);
\draw[arrow] (rerank) -- (gen);
\draw[arrow] (gen) -- (metrics);
\draw[arrow, dashed] (rrf) to[bend right=18] node[below, align=center] {no-rerank\\ablation} (gen);

\node[draw, dashed, rounded corners=2pt, fit=(bm25)(dense)(rrf)(rerank), inner sep=6pt, label={[font=\small]above:Retrieval and reranking}] {};
\end{tikzpicture}}
\caption{SciRet pipeline evaluated in this paper. Text is chunked once and indexed for both sparse and dense retrieval. Ranked lists are fused with reciprocal rank fusion; an off-the-shelf cross-encoder reranker is evaluated as an ablation before grounded generation and evaluation. The diagram is rendered as vector graphics to avoid raster blurring in the submission PDF.}
\label{fig:pipeline}
\end{figure*}
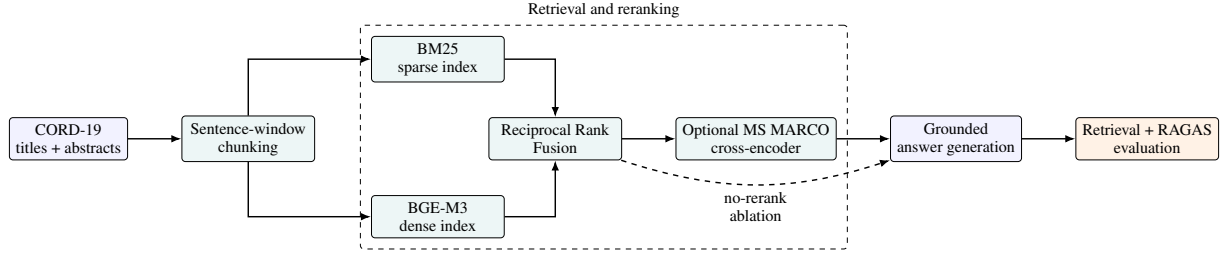

Our goal is not to claim a new state-of-the-art architecture. Instead, we report a compact, reproducible study of practical design choices for resource-constrained scientific RAG: BM25 versus dense retrieval, sparse-dense fusion, off-the-shelf cross-encoder reranking, and automated generation evaluation. Figure~\ref{fig:pipeline} summarizes the system. This is a focused empirical study with 15 evaluation queries per scale, intended to compare system behavior under controlled conditions rather than establish a benchmark.

\paragraph{Contributions.}
We make three contributions:
(1) a controlled multi-scale evaluation of a fixed scientific RAG pipeline across 1K, 5K, and 15K CORD-19 paper samples;
(2) an empirical analysis showing that BM25+BGE-M3 fusion is more robust than either component alone in our setting; and
(3) a negative reranking result: an MS MARCO-trained cross-encoder reduces precision on scientific text, highlighting domain mismatch as a practical risk.

\section{Related Work}

\paragraph{Scientific RAG and QA.}
RAG combines retrieval with generation for knowledge-intensive NLP \citep{lewis2020rag}. Scientific and biomedical QA benchmarks such as BioASQ \citep{bioasq}, SciFact \citep{wadden2020scifact}, COVID-QA \citep{moller2020covidqa}, and CORD-19 \citep{wang2020cord19} emphasize that retrieval quality and evidence grounding are central to scientific question answering. SciRet focuses on the retrieval and reranking behavior of a practical scientific RAG pipeline rather than on training a new generator.

\paragraph{Sparse, dense, and hybrid retrieval.}
BM25 remains a strong lexical baseline for scientific text because exact terms, abbreviations, and biomedical entities matter \citep{robertson2009bm25}. Dense retrievers such as DPR \citep{karpukhin2020dpr}, Sentence-BERT \citep{reimers2019sbert}, SPECTER \citep{cohan2020specter}, and BGE-M3 \citep{chen2024bgem3} support semantic matching beyond lexical overlap. Hybrid retrieval is often effective because sparse and dense systems retrieve complementary evidence; reciprocal rank fusion (RRF) is a simple and robust fusion method \citep{cormack2009rrf}. BEIR further shows that retrieval behavior can vary substantially across domains and datasets \citep{thakur2021beir}.

\paragraph{Reranking and evaluation.}
Cross-encoders can improve ranking by jointly encoding query-passage pairs, but transfer depends on training data. We test an MS MARCO-trained reranker in a scientific setting and find that it hurts precision. For generation, we use RAGAS \citep{es2023ragas} as an automated comparative signal, while recognizing that LLM-based evaluation can inherit judge-model biases. Claim-level factuality metrics such as FActScore \citep{min2023factscore} motivate future citation-integrity evaluation.

\section{Method}

\subsection{Corpus and Scale Protocol}

We evaluate on CORD-19 \citep{wang2020cord19}. To keep the study compute-aware and reproducible, we index titles and abstracts rather than full text. This reduces storage and embedding cost, but limits evidence depth; we treat this as a limitation rather than a complete scientific-document solution.

We construct three corpus scales while keeping all settings fixed: 1K papers, 5K papers, and 15K papers. The resulting chunk counts are 1,034, 5,160, and 15,480, respectively (Table~\ref{tab:dataset}). All experiments use the same sentence-window chunking strategy, embedding model, retrieval cutoffs, RRF parameter, generation prompt style, and evaluation scripts.

\begin{table}[t]
\centering
\small
\caption{Dataset statistics for the three evaluation scales.}
\label{tab:dataset}
\begin{tabular}{lrrr}
\toprule
Statistic & 1K & 5K & 15K \\
\midrule
Papers indexed & 1,000 & 5,000 & 15,000 \\
Text chunks & 1,034 & 5,160 & 15,480 \\
Mean chunk tokens & 215 & 215 & 215 \\
Evaluation queries & 15 & 15 & 15 \\
\bottomrule
\end{tabular}
\end{table}

\subsection{Retrieval and Reranking}

We compare three retrieval systems:
\textbf{Dense}: BGE-M3 embeddings with vector search;
\textbf{BM25}: sparse lexical retrieval;
and \textbf{Hybrid}: reciprocal rank fusion of the dense and BM25 ranked lists.
RRF scores a document $d$ as:
\begin{equation}
  \mathrm{RRF}(d)=\sum_i \frac{1}{60+\mathrm{rank}_i(d)}.
\end{equation}
Stage 1 retrieves 50 candidates from each retrieval system before fusion.

We then evaluate an optional cross-encoder reranker, \texttt{cross-encoder/ms-marco-\allowbreak{}MiniLM-L-6-v2}. Because this reranker is trained on web-search data, its performance on scientific abstracts is an empirical question rather than a guaranteed improvement.

\subsection{Generation and Metrics}

Retrieved passages are assembled into a grounded prompt for GPT-4o-mini, which is instructed to answer only from retrieved context and cite passages by index. We report retrieval Recall@$K$ for $K \in \{1,3,5,10,20\}$ and Precision@$K$ for reranking ablations. Generation is evaluated with RAGAS faithfulness, answer relevancy, context precision, and context recall.

\subsection{Retrieval Evaluation Note}
\label{sec:evalnote}

Retrieval evaluation uses pseudo-relevance labels: the top-3 hybrid results per query are treated as relevant. This makes our results useful for controlled comparison and debugging, but it introduces circularity that may favor the hybrid system, and a 15-query set limits statistical power. All retrieval figures should therefore be read as controlled comparative evidence rather than benchmark scores. We return to this limitation in Section~\ref{sec:limitations}.

\section{Results}

\subsection{Hybrid Retrieval Is Most Robust}

\begin{figure}[t]
  \centering
  \includegraphics[width=\linewidth]{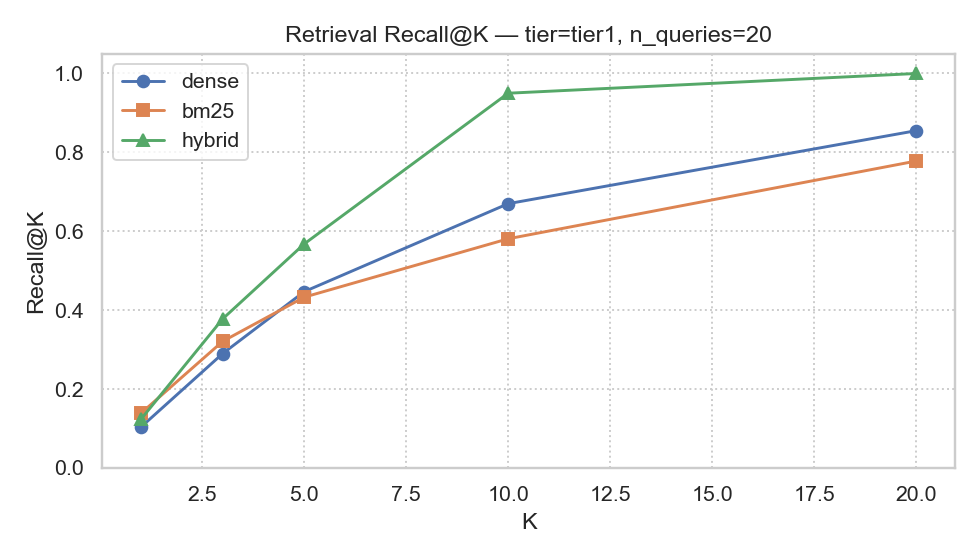}
  \caption{Recall curves at the 1K scale. Hybrid retrieval improves over either component alone at larger cutoffs.}
  \label{fig:recall_curve}
\end{figure}

Table~\ref{tab:recall} reports Recall@$K$ across scales. Hybrid retrieval dominates at $K \geq 5$ and reaches Recall@10 of 1.000 at 1K and 15K. Dense retrieval is competitive at a small scale, while BM25 remains competitive at R@1, especially at 15K. Figure~\ref{fig:recall_curve} visualizes the recall curves at the 1K scale.

\begin{table}[!t]
\centering
\scriptsize
\caption{Recall@$K$ across scales. With three pseudo-relevant documents per query, the maximum possible R@1 is $1/3=0.333$. Best value per scale and cutoff is bolded.}
\label{tab:recall}
\setlength{\tabcolsep}{2.5pt}
\begin{tabular}{llccccc}
\toprule
Scale & System & R@1 & R@3 & R@5 & R@10 & R@20 \\
\midrule
\multirow{3}{*}{1K}
& Dense & \textbf{0.333} & \textbf{0.720} & 0.720 & 0.767 & 0.793 \\
& BM25 & \textbf{0.333} & 0.513 & 0.573 & 0.667 & 0.693 \\
& Hybrid & 0.267 & 0.627 & \textbf{0.820} & \textbf{1.000} & \textbf{1.000} \\
\midrule
\multirow{3}{*}{5K}
& Dense & \textbf{0.333} & \textbf{0.733} & 0.740 & 0.807 & 0.827 \\
& BM25 & 0.327 & 0.513 & 0.573 & 0.627 & 0.687 \\
& Hybrid & 0.260 & 0.613 & \textbf{0.847} & \textbf{0.993} & \textbf{1.000} \\
\midrule
\multirow{3}{*}{15K}
& Dense & 0.147 & 0.400 & 0.560 & 0.787 & 0.920 \\
& BM25 & 0.180 & 0.387 & 0.560 & 0.747 & 0.847 \\
& Hybrid & \textbf{0.333} & \textbf{1.000} & \textbf{1.000} & \textbf{1.000} & \textbf{1.000} \\
\bottomrule
\end{tabular}
\end{table}

\subsection{Generic Reranking Hurts Precision}

Table~\ref{tab:rerank} shows that the MS MARCO-trained cross-encoder reduces precision at all reported cutoffs. At 1K, P@5 drops from 0.600 to 0.404. At 5K, it drops from 0.600 to 0.368. Since the no-rerank baseline is identical at both scales, the degradation is attributable to reranking rather than to weaker Stage 1 retrieval.

\begin{table}[!t]
\centering
\small
\caption{Precision@$K$ before and after cross-encoder reranking.}
\label{tab:rerank}
\begin{tabular}{llcccc}
\toprule
Scale & System & P@1 & P@3 & P@5 & P@10 \\
\midrule
\multirow{2}{*}{1K}
& No rerank & 1.000 & 1.000 & 0.600 & 0.300 \\
& Reranked & 0.680 & 0.527 & 0.404 & 0.234 \\
\midrule
\multirow{2}{*}{5K}
& No rerank & 1.000 & 1.000 & 0.600 & 0.300 \\
& Reranked & 0.620 & 0.467 & 0.368 & 0.250 \\
\bottomrule
\end{tabular}
\end{table}

\subsection{Generation Scores Improve With Scale}

Table~\ref{tab:ragas} reports RAGAS generation metrics. Faithfulness increases from 0.917 at 1K to 0.960 at 15K, and answer relevancy increases from 0.680 to 0.870. Context precision remains low, indicating that retrieved contexts include topic-related but not always tightly targeted passages.

\begin{table}[!t]
\centering
\small
\caption{RAGAS generation quality across scales.}
\label{tab:ragas}
\begin{tabular}{lcccc}
\toprule
Scale & Faith. & Ans. Rel. & Ctx. Prec. & Ctx. Rec. \\
\midrule
1K & 0.917 & 0.680 & 0.095 & 0.260 \\
5K & 0.940 & 0.703 & 0.108 & 0.240 \\
15K & 0.960 & 0.870 & 0.122 & 0.100 \\
\bottomrule
\end{tabular}
\end{table}

\FloatBarrier

\section{Discussion}

The results support two practical lessons. First, sparse-dense fusion is a strong default for scientific RAG under limited compute: BM25 and BGE-M3 retrieve complementary evidence, and RRF is simple enough to run across scales. Second, reranking is not automatically beneficial. The MS MARCO cross-encoder harms ranking on our scientific corpus, suggesting that domain-adapted reranking should be tested before deployment.

The study also shows why compute-aware evaluation matters. The 1K scale is useful for debugging, but some retrieval behavior changes at 15K. Small-scale experiments should therefore be treated as development checks rather than final evidence.

\section{Limitations}
\label{sec:limitations}

This paper has important limitations. First, we index only titles and abstracts, not full papers, figures, or tables. Second, retrieval evaluation uses pseudo-relevance labels derived from hybrid top-3 results, which can favor hybrid retrieval and cannot replace independent relevance annotation. Third, the evaluation set contains 15 queries, limiting statistical power. Fourth, RAGAS provides an automated signal but not a substitute for human or claim-level citation verification. We frame SciRet as a reproducible empirical study and a basis for stronger follow-up work, not as a complete scientific QA benchmark.

\section{Conclusion}

We presented SciRet, a compute-aware empirical study of retrieval and reranking for scientific RAG over CORD-19. Across 1K, 5K, and 15K paper samples, BM25+BGE-M3 fusion is more robust than either component alone in our setting, while an MS MARCO-trained cross-encoder reranker consistently reduces precision. These results suggest that standard RAG components should be tested carefully in scientific QA rather than assumed to transfer from web search settings. Hybrid retrieval appears to be a strong default under limited compute, but the retrieval evaluation here is based on pseudo-relevance labels and a small query set, so the findings should be read as controlled empirical evidence rather than final benchmark results.Future work should prioritize independent relevance annotation to remove the pseudo-label circularity, full-document evidence beyond titles and abstracts, domain-adapted rerankers trained on scientific text, and claim-level citation verification to ground generated answers in verifiable evidence.

\section*{Ethics Statement}

SciRet is evaluated on scientific literature and does not introduce new human-subject data. Because scientific QA systems can influence user interpretation of biomedical evidence, generated answers should not be used for medical decision-making without expert review. The current system uses title and abstract evidence only and may omit important full-text context.

\paragraph{Generative AI Disclosure.}
The authors utilized Claude and ChatGPT to assist in refining prose, improving readability, and copyediting select portions of this manuscript. The core ideas, technical methodology, and experimental evaluations were fully developed by the authors. The authors assume full responsibility for the final content.
\section*{Acknowledgments}


\bibliography{custom}

@article{chen2024bgem3,
  author  = {Jianlv Chen and Shitao Xiao and Peitian Zhang and Kun Luo and Defu Lian and Zheng Liu},
  title   = {{BGE M3-Embedding}: Multi-Lingual, Multi-Functionality, Multi-Granularity Text Embeddings Through Self-Knowledge Distillation},
  journal = {arXiv preprint arXiv:2402.03216},
  year    = {2024}
}

@inproceedings{cohan2020specter,
  author    = {Arman Cohan and Sergey Feldman and Iz Beltagy and Doug Downey and Daniel S. Weld},
  title     = {{SPECTER}: Document-Level Representation Learning Using Citation-Informed Transformers},
  booktitle = {Proceedings of the 58th Annual Meeting of the Association for Computational Linguistics},
  pages     = {2270--2282},
  year      = {2020}
}

@inproceedings{cormack2009rrf,
  author    = {Gordon V. Cormack and Charles L. A. Clarke and Stefan Buettcher},
  title     = {Reciprocal Rank Fusion Outperforms {Condorcet} and Individual Rank Learning Methods},
  booktitle = {Proceedings of the 32nd International ACM SIGIR Conference on Research and Development in Information Retrieval},
  pages     = {758--759},
  year      = {2009}
}

@article{es2023ragas,
  author  = {Shahul Es and Jithin James and Luis Espinosa-Anke and Steven Schockaert},
  title   = {{RAGAS}: Automated Evaluation of Retrieval Augmented Generation},
  journal = {arXiv preprint arXiv:2309.15217},
  year    = {2023}
}

@inproceedings{karpukhin2020dpr,
  author    = {Vladimir Karpukhin and Barlas O{\u{g}}uz and Sewon Min and Patrick Lewis and Ledell Wu and Sergey Edunov and Danqi Chen and Wen{-}tau Yih},
  title     = {Dense Passage Retrieval for Open-Domain Question Answering},
  booktitle = {Proceedings of the 2020 Conference on Empirical Methods in Natural Language Processing},
  pages     = {6769--6781},
  year      = {2020}
}

@inproceedings{lewis2020rag,
  author    = {Patrick Lewis and Ethan Perez and Aleksandra Piktus and Fabio Petroni and Vladimir Karpukhin and Naman Goyal and Heinrich K{\"u}ttler and Mike Lewis and Wen{-}tau Yih and Tim Rockt{\"a}schel and Sebastian Riedel and Douwe Kiela},
  title     = {Retrieval-Augmented Generation for Knowledge-Intensive {NLP} Tasks},
  booktitle = {Advances in Neural Information Processing Systems},
  volume    = {33},
  pages     = {9459--9474},
  year      = {2020}
}

@inproceedings{min2023factscore,
  author    = {Sewon Min and Kalpesh Krishna and Xinxi Lyu and Mike Lewis and Wen{-}tau Yih and Pang Wei Koh and Mohit Iyyer and Luke Zettlemoyer and Hannaneh Hajishirzi},
  title     = {{FActScore}: Fine-Grained Atomic Evaluation of Factual Precision in Long Form Text Generation},
  booktitle = {Proceedings of the 2023 Conference on Empirical Methods in Natural Language Processing},
  pages     = {12076--12100},
  year      = {2023}
}

@inproceedings{moller2020covidqa,
  author    = {Timo M{\"o}ller and Anthony Reina and Raghavan Jayakumar and Malte Pietsch},
  title     = {{COVID-QA}: A Question Answering Dataset for {COVID-19}},
  booktitle = {Proceedings of the 1st Workshop on NLP for COVID-19 at ACL 2020},
  year      = {2020}
}

@inproceedings{reimers2019sbert,
  author    = {Nils Reimers and Iryna Gurevych},
  title     = {Sentence-{BERT}: Sentence Embeddings Using Siamese {BERT}-Networks},
  booktitle = {Proceedings of the 2019 Conference on Empirical Methods in Natural Language Processing and the 9th International Joint Conference on Natural Language Processing},
  pages     = {3982--3992},
  year      = {2019}
}

@article{robertson2009bm25,
  author  = {Stephen Robertson and Hugo Zaragoza},
  title   = {The Probabilistic Relevance Framework: {BM25} and Beyond},
  journal = {Foundations and Trends in Information Retrieval},
  volume  = {3},
  number  = {4},
  pages   = {333--389},
  year    = {2009}
}

@inproceedings{thakur2021beir,
  author    = {Nandan Thakur and Nils Reimers and Andreas R{\"u}ckl{\'e} and Abhishek Srivastava and Iryna Gurevych},
  title     = {{BEIR}: A Heterogeneous Benchmark for Zero-Shot Evaluation of Information Retrieval Models},
  booktitle = {Proceedings of the Thirty-Fifth Conference on Neural Information Processing Systems Datasets and Benchmarks Track},
  year      = {2021}
}

@article{bioasq,
  author  = {George Tsatsaronis and Georgios Balikas and Prodromos Malakasiotis and Ioannis Partalas and Matthias Zschunke and Michael R. Alvers and Dirk Weissenborn and Anastasia Krithara and Sergios Petridis and Dimitris Polychronopoulos},
  title   = {An Overview of the {BioASQ} Large-Scale Biomedical Semantic Indexing and Question Answering Competition},
  journal = {BMC Bioinformatics},
  volume  = {16},
  pages   = {138},
  year    = {2015}
}

@inproceedings{wadden2020scifact,
  author    = {David Wadden and Shanchuan Lin and Kyle Lo and Lucy Lu Wang and Madeleine van Zuylen and Arman Cohan and Hannaneh Hajishirzi},
  title     = {Fact or Fiction: Verifying Scientific Claims},
  booktitle = {Proceedings of the 2020 Conference on Empirical Methods in Natural Language Processing},
  pages     = {7534--7550},
  year      = {2020}
}

@article{wang2020cord19,
  author  = {Lucy Lu Wang and Kyle Lo and Yoganand Chandrasekhar and Russell Reas and Jiangjiang Yang and Doug Baber and Kathryn Eide and Brendan Ros and Nascence Kim and Spencer Wilhelm},
  title   = {{CORD-19}: The {COVID-19} Open Research Dataset},
  journal = {arXiv preprint arXiv:2004.10706},
  year    = {2020}
}

\appendix

\section{Appendix Overview}

The appendix preserves supporting evidence that is useful for review but not essential to the short-paper narrative: chunking analysis, retrieval complementarity, embedding sanity checks, question-set details, compute budget, and supplementary visualizations.

\section{Chunking Analysis}

We compared fixed-character, fixed-token, sentence-window, and paragraph-based chunking at the 1K scale. Sentence-window chunking achieved the strongest boundary cleanliness and was used across all scales.

\begin{table}[H]
\centering
\small
\caption{Chunking strategy comparison on the 1K corpus.}
\label{tab:chunking}
\begin{tabular}{lrrr}
\toprule
Strategy & Chunks & Tokens & Clean \% \\
\midrule
Fixed character & 629 & 182.4 & 69.2 \\
Fixed token & 535 & 210.3 & 81.9 \\
Sentence window & 519 & 215.8 & 85.4 \\
Paragraph & 500 & 222.0 & 84.8 \\
\bottomrule
\end{tabular}
\end{table}

\begin{figure}[H]
  \centering
  \includegraphics[width=.8\linewidth]{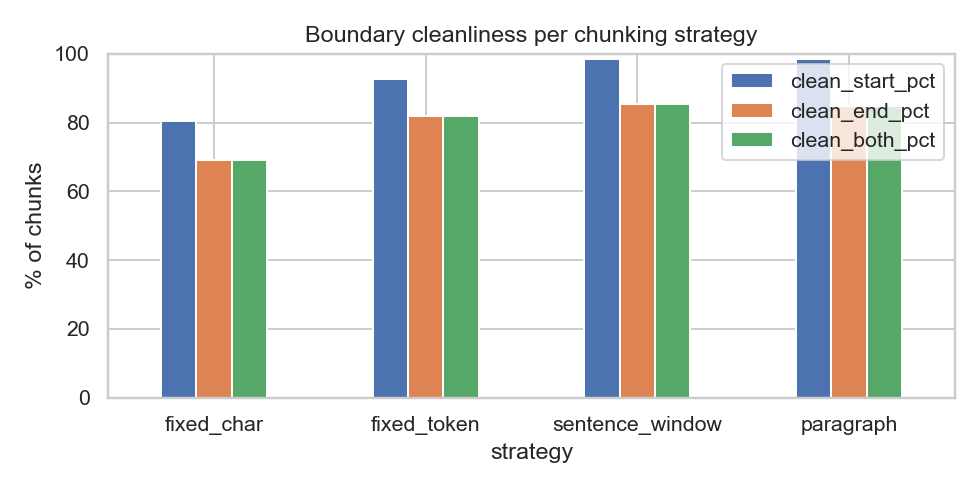}
  \caption{Boundary cleanliness by chunking strategy.}
\end{figure}

\section{Retrieval Complementarity}

At the 1K scale, the mean Jaccard similarity between dense and BM25 top-10 result sets is 0.213, meaning roughly 79\% of retrieved documents differ between the two methods. This supports the use of rank fusion.

\begin{figure}[H]
  \centering
  \includegraphics[width=.85\linewidth]{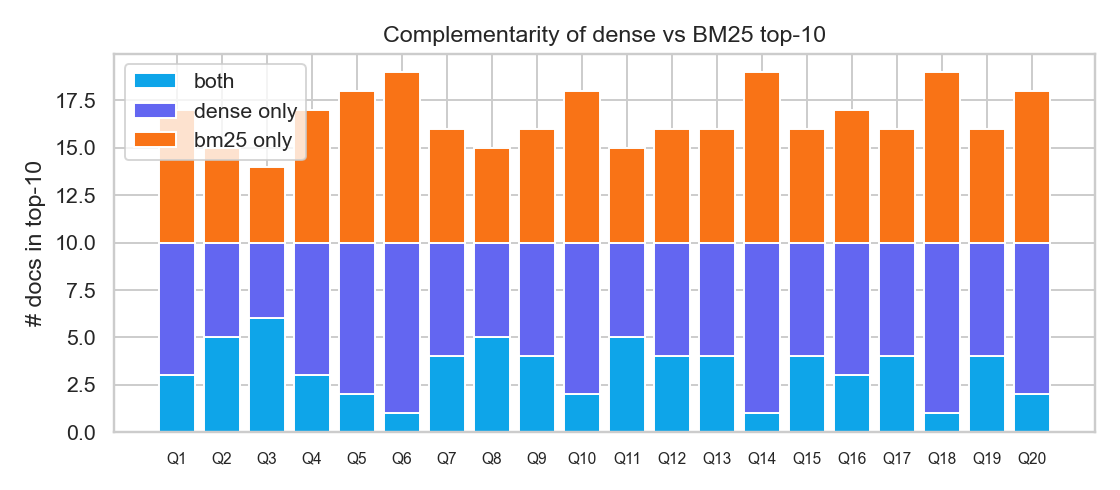}
  \caption{Overlap between dense and BM25 top-10 retrieval results across queries.}
\end{figure}

\section{Embedding Sanity Check}

BGE-M3 related-passage similarities exceed unrelated-passage similarities by 0.16 to 0.41 cosine units across biomedical term, symptom, and treatment queries.

\begin{table}[H]
\centering
\small
\caption{Cosine similarity sanity check for BGE-M3 embeddings.}
\begin{tabular}{lccc}
\toprule
Query type & Related & Unrelated & Margin \\
\midrule
Biomedical term & 0.81 & 0.65 & +0.16 \\
Symptom-based & 0.79 & 0.44 & +0.35 \\
Treatment & 0.83 & 0.42 & +0.41 \\
\bottomrule
\end{tabular}
\end{table}

\section{Reranker Rank Movement}

\begin{figure}[H]
  \centering
  \includegraphics[width=.8\linewidth]{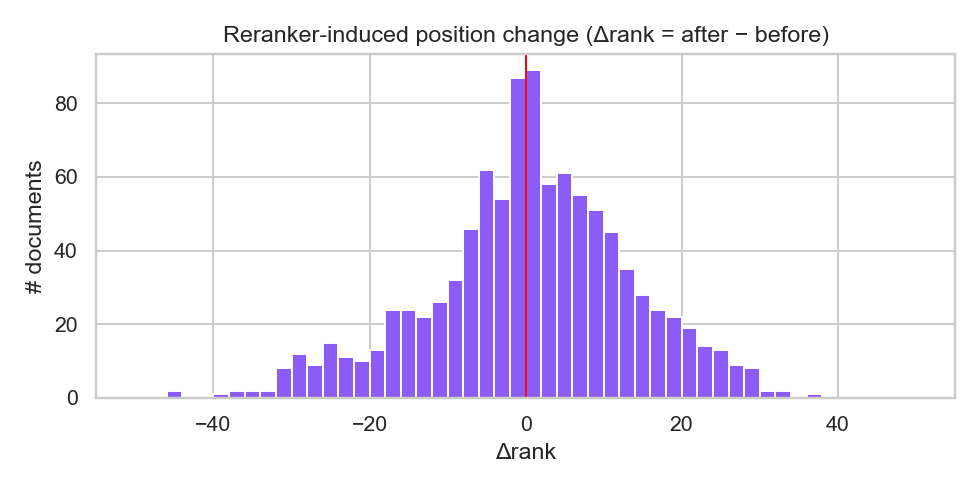}
  \caption{Distribution of reranker-induced rank shifts at the 1K scale.}
\end{figure}

\section{Evaluation Questions}

The 15 evaluation questions cover clinical, epidemiological, and treatment themes in CORD-19. Examples include: ``What are the primary transmission routes of SARS-CoV-2?'', ``Which comorbidities increase COVID-19 mortality risk?'', and ``What is the efficacy of remdesivir in treating COVID-19?'' We release the question set, pseudo-relevance labels, and raw result files with the project.

\section{Compute Budget}

\begin{table}[H]
\centering
\small
\caption{Approximate wall-clock time by pipeline stage.}
\begin{tabular}{lrrr}
\toprule
Stage & 1K & 5K & 15K \\
\midrule
Chunking & $<$1 min & 3 min & 8 min \\
Embedding & 18 min & 25 min & 73 min \\
BM25 index & $<$1 min & $<$1 min & 2 min \\
Retrieval & $<$1 min & $<$1 min & $<$1 min \\
Reranking & 4 min & 12 min & N/A \\
Generation & 8 min & 8 min & 8 min \\
\bottomrule
\end{tabular}
\end{table}

\end{document}